# Self-interpretable Convolutional Neural Networks for Text Classification

Wei Zhao[1], Rahul Singh, Tarun Joshi, Agus Sudjianto, and Vijayan N. Nair

Corporate Model Risk, Wells Fargo, USA

May 14, 2021

**Abstract**

Deep learning models for natural language processing (NLP) are inherently complex and often viewed as black box in nature. This paper develops an approach for interpreting convolutional neural networks for text classification problems by exploiting the local-linear models inherent in ReLU-DNNs. The CNN model combines the word embedding through convolutional layers, filters them using max-pooling, and optimizes using a ReLU-DNN for classification. To get an overall self-interpretable model, the system of local linear models from the ReLU DNN are mapped back through the max-pool filter to the appropriate n-grams. Our results on experimental datasets demonstrate that our proposed technique produce parsimonious models that are self-interpretable and have comparable performance with respect to a more complex CNN model. We also study the impact of the complexity of the convolutional layers and the classification layers on the model performance.

## 1. Introduction

Deep neural networks (DNNs) have become the dominant approach in natural language processing (NLP) and have been used for solving real world problems related to text such as sentiment analysis, machine translation, summarization and topic categorization. Text classification is an important task in many of these applications. For example, DNNs have been used for categorization of complaints, emails, etc. in banking, while they are applied widely to classify customer reviews into positive or negative reviews in online shopping.

While DNNs have excellent predictive performance, the models are complex. In regulated industries such as banking, models have to be interpretable so that the results can be easily explained to various stakeholders including customers and regulators. From this perspective, simple models are still preferred as they are intrinsically interpretable. However, simple models suffer from poor predictive performance as they o capture complex interactions among input features that occur frequently. Hence, there is a trade-off between model performance and model interpretability. This motivates us to seek self-interpretable models with good predictive performance. This paper proposes an approach for NLP text classification tasks.

---

[1] Email: vivienzhao0119@hotmail.com

There are several techniques in the literature for interpretability of ML algorithms. These include post hoc techniques such as LIME (Ribeiro, Singh, & Guestrin, 2016) which bases the explanation on linear surrogate models, and kernel SHAP (Lundberg & Lee, 2017; Lundberg, Erion, & Lee, 2018) which computes a local importance value for each model feature. Others include gradient-based diagnostics like Integrated Gradients (Sundararajan, Taly, & Yan, 2017) that have the advantage of fast computation. However, for explaining DNNs that are intrinsically complex, most of these techniques require assumptions and approximations and hence can potentially lead to inaccurate model explanation.

There have also been attempts to extract feature importance in DNN models by analyzing their inner workings. Examples include calculating feature importance in CNN networks by studying the pooling layer (Jacovi, Shalom, & Goldberg, 2018), and layer-wise relevance propagation (Bach, et al., 2015) (Arras, Horn, Müller, & Samek, 2017). (Zhao, Joshi, Nair, & Sudjianto, 2020) use the max-pooling concept in (Jacovi, Shalom, & Goldberg, 2018) to extract features in the convolutional layers and utilize SHAP to generate local explainability in text classification models. In this paper, we use a similar convolutional framework as in (Zhao, Joshi, Nair, & Sudjianto, 2020) but replace SHAP with local linear models implicit in ReLU networks for interpretation. Our ideas are based on a general framework for intrinsic interpretability of ReLU DNN networks in (Sudjianto, Knauth, Singh, Yang, & Zhang, 2020).

The CNN structure has been shown to have good performances in these tasks (Kim, 2014). The convolutional layer does feature engineering, which selects important information from the text documents, and is interpretable when used in conjunction with max-pooling layer. We apply regularization on the ReLU DNN after the convolutional layer to reduce model complexity. Our experiments demonstrate such regularized model can achieve very good predictive performances and are simpler for model interpretation.

The paper is organized as follows. In Section 2, we discuss the structure of CNN model that is used in NLP text classification and the methods proposed in Aletheia (Sudjianto, Knauth, Singh, Yang, & Zhang, 2020) for interpreting ReLU DNN classification layer. In Section 3, we propose a technique to build simple and well performing text classification models based on the CNN structure. In addition, we introduce the steps for model interpretation by assigning filter importance achieved from ReLU DNN to extracted text features. In Section 4, we apply this technique on Yelp Reviews dataset to demonstrate our results. We summarize our findings and conclusions in Section 5.

## 2. Background

This section reviews the key components that are need for our approach: i) the convolutional layer structure in CNN text classification discussed in (Zhao, Joshi, Nair, & Sudjianto, 2020) ; and ii) the ReLU-based DNN for self-explanatory local linear models proposed by (Sudjianto, Knauth, Singh, Yang, & Zhang, 2020); iii) the Aletheia merging algorithm also proposed by (Sudjianto,

Knauth, Singh, Yang, & Zhang, 2020). Figure 1 shows the complete CNN model structure along with the first two components.

## 2.1 Structure of Convolutional Layer

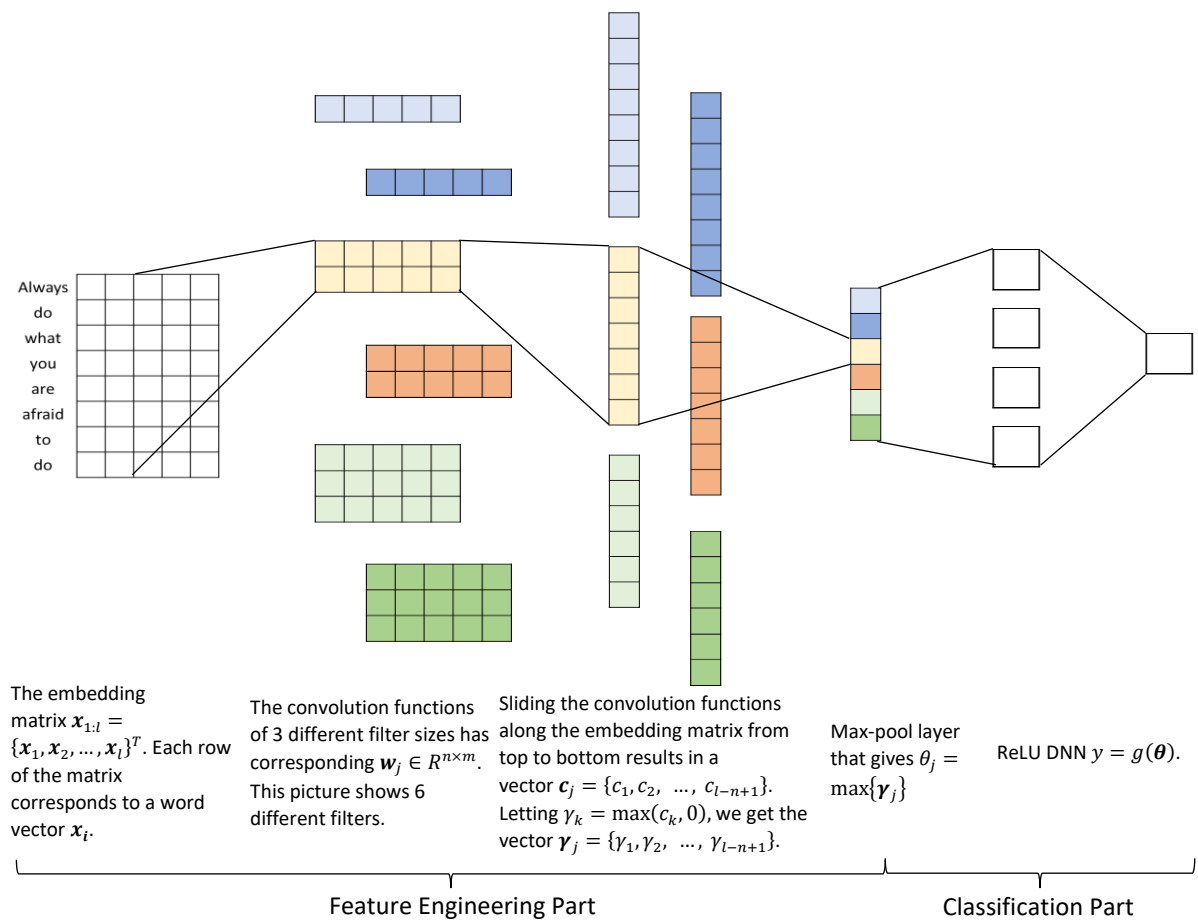

The embedding matrix $x_{1:l} = \{x_1, x_2, \ldots, x_l\}^T$. Each row of the matrix corresponds to a word vector $x_i$.

The convolution functions of 3 different filter sizes has corresponding $w_j \in R^{n \times m}$. This picture shows 6 different filters.

Sliding the convolution functions along the embedding matrix from top to bottom results in a vector $c_j = \{c_1, c_2, \ldots, c_{l-n+1}\}$. Letting $\gamma_k = \max(c_k, 0)$, we get the vector $\gamma_j = \{\gamma_1, \gamma_2, \ldots, \gamma_{l-n+1}\}$.

Max-pool layer that gives $\theta_j = \max\{\gamma_j\}$

ReLU DNN $y = g(\theta)$.

Feature Engineering Part          Classification Part

*Figure 1: CNN text classification model divided into feature engineering part and classification part*

In NLP tasks, the unstructured input text document with $l$ words is usually transformed first to word embedding: a $l \times m$ matrix that embeds each word/token into an $m$-dimensional vector space

$$x_{1:l} = \{x_1, x_2, \ldots, x_l\}^T \in R^{l \times m}.$$

Then we apply a size $n$ filter consisting of convolutional function $f_j \in R^{n \times m} \to R, j \in \{1, 2, \ldots, h\}$ followed by a max-pool layer. When $f_j$ slides over the embedding matrix, the filter selects an $n$-gram text feature from the input embedding matrix:

$$\theta_j = \max\{f_j(x_{1:n}), f_j(x_{2:n-1}), \ldots, f_j(x_{l-n+1:l}), 0\},$$

where $f_j(x_{k:k+n-1}) = \boldsymbol{w}_j \cdot x_{k:k+n-1} + b_j$, and $k \in \{1,2,\ldots,l-n+1\}$, and $\boldsymbol{w}_j \in R^{n \times m}$. With $h$ different filters scanning the embedding inputs, the model is able to select $h$ different text features ($n$-grams) from the original text documents:

$$\boldsymbol{\theta} = \{\theta_1, \theta_2, \ldots, \theta_h\}.$$

The $\boldsymbol{\theta}$ here is called a max-pooling layer in the CNN structure. It selects the maximum value outputted from each convolution function. Each selected maximum value can be mapped back to a specific submatrix in the embedding matrix input, which corresponds to an n-gram from the text.

A predictive model is then fit to the results of the max-pool layer are to get a class prediction score for final classification decision:

$$y = g(\boldsymbol{\theta}).$$

In our application, we will use a ReLU DNN (described in the next section) as it leads to self-explanatory local linear models. Note that the feature extraction step is trained together with the ReLU DNN classifier. Therefore, different classifier structures (i.e. the number of hidden neurons and the number of hidden layers) can result in different features extracted by convolutional filters. The details will be illustrated in Section 4 using real data.

## 2.2 ReLU DNN and Self-Explanatory Local-Linear Models

It is known (see, for example, (Sudjianto, Knauth, Singh, Yang, & Zhang, 2020)) that a ReLU DNN partitions the input space into multiple sub-regions, and each sub-region is determined by a unique activation pattern. The fitted predictors within each region are determined by a local-linear model (LLM). Specifically, by multi-layer propagation, the original input features are sequentially transformed by

$$z^{(l)} = W^{(l-1)}\chi^{(l-1)} + b^{(l-1)}, \quad \text{for } l = 1, \ldots L \text{ and } \chi^{(0)} = \boldsymbol{\theta}.$$

Then, the ReLU activation function is applied as shown below,

$$\chi_i^{(l)} = max\{0, z_i^{(l)}\}, \quad \text{for } i = 1, \ldots n_l \text{ and } l = 1, \ldots L.$$

Finally, the prediction takes the linear form:

$$\eta(x) = W^{(L)}\chi^{(L)} + b^{(L)}.$$

A linear function over linear function is still linear, so there exists a final $W^\chi$ and a final $b^\chi$ such that

$$\eta(x) = W^\chi \chi^{(0)} + b^\chi. \tag{1}$$

We can use the LLMs to perform model diagnostics as the linear representation facilitates local interpretation of the model and can be used to provide region-wise explanations (see Figure 2).

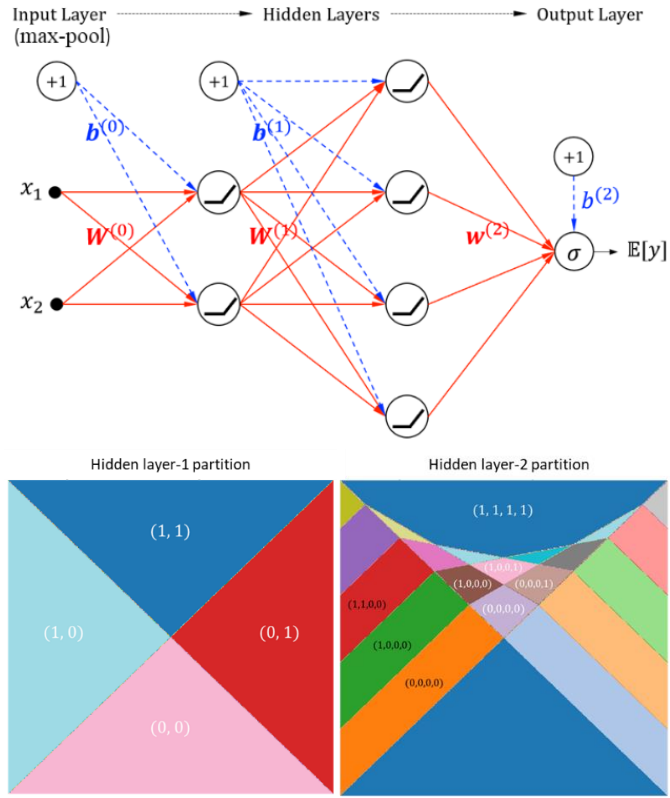

*Figure 2. A ReLU DNN showing forward propagation after the max-pooling layer in a CNN model and the resulting data partitions*

### 2.3 Merging (Alethiea)

A direct application of the usual ReLU DNN algorithm can result in a large number of regions. Further simplification of the network can be done by merging several homogeneous regions into a single one. See Aletheia (Sudjianto, Knauth, Singh, Yang, & Zhang, 2020) for details.

In this paper, we accomplish this in two steps: a) use Agglomerative hierarchal clustering to combine LLMs with similar coefficients; and b) use nearest neighbor clustering to remove remaining small regions. Agglomerative clustering combines LLMs with similar coefficients, where similarity is measured by Euclidean distance between the coefficients, subject to the connectivity constraint. Connectivity constraint guarantees that only spatially close regions are merged. However, this step sometimes still results in regions that have relatively small number of data points. Hence, another step is performed in which smaller regions are combined with neighboring large regions. For more details, see (Sudjianto, Knauth, Singh, Yang, & Zhang, 2020).

## 3. Methodology

This section combines the ideas in Section 2 to propose a novel technique for developing self-interpretable Convolutional Neural Network (CNN)-based models for text classification. The proposed approach consists of convolutional filters followed by a ReLU DNN classifier. We use

L1/L2 regularization over the ReLU DNN to simplify the model while maintaining comparable model performances. We then apply Aletheia (Sudjianto, Knauth, Singh, Yang, & Zhang, 2020) on the ReLU DNN layers and provide interpretations for the extracted features which can be mapped to n-grams in the input text.

The following terms will be referenced in the rest of the paper:

- Filter: convolutional layer + activation function + max-pool layer
- Filter size: the kernel size of the 1D convolution, which directly determines the size of n-grams fed into the filter.
- Extracted feature: the max-pool layer outputs.
- Feature engineering layer: the convolutional layers that produce interpretable text features.
- ReLU Classifier: the feed forward layers constructed over the max-pool layer outputs with ReLU activation.

The complete approach is described below and consists of the following three steps: i) CNN text classification model building; ii) ReLU DNN classifier simplification; iii) and CNN text classification model interpretation.

**Model Building Step**: The CNN structure described in Section 2.1 Structure of ) is used for building the text classification model in the first step. We use three different filter sizes for uni-gram, bi-gram and tri-gram features. We use $m$ different filters for each size. In other words, the model will finally select $3 \times m$ text features ($n$-grams) from the original text document. Ideally, the filters should extract different semantic features from the text. However, in practice, there is some duplication or overlapping of $n$-grams.

The ReLU classifier is constructed by two linear layers with ReLU activation. The output of the first layer has $k$ hidden neurons, which are passed into the second layer. The second layer is the final layer and reduces the $k$-dimensional input into dimension of one for binary classification or dimension $K$ for K-class multi-classification problems.

**Model Simplification Step**: We also study model simplification using L1 regularization on the parameters on the linear layers of the ReLU classifier. As explained in Section 2.2 ReLU DNN and Self-Explanatory Local-Linear Models ), the ReLU classifier is equivalent to a set of local linear models (LLMs). In the case with one hidden layer with $k$ hidden neurons, the theoretical maximum number of LLMs we end up getting are $2^k$. If $k$ is too large, there will be many LLMs and the interpretation of the model becomes cumbersome. In fact, even with 10 hidden neurons, we can still have more than hundred LLMs to interpret. This motivates us to apply L1 regularization over the ReLU classifier. We will show empirically in our experiments that adding L1 regularization on the hidden neurons drastically reduces the total number of LLMs, and at the same time does not reduce the prediction performance compared to the original model. If there still redundant LLMs, we apply the merging algorithm in Aletheia after the model is trained. The algorithm merges multiple local linear regions using clustering methods and refits linear models at final linear regions.

**Model Interpretation Step**: The final step is self-interpretation of the CNN text model. We first interpret the ReLU classifier by looking at the local weights $W$ of the LLM in each sub-region. We define *importance of a text feature i* as being proportional to the magnitude of the local weight $w_i$. For example, if a particular text sample $\theta$ falls into region $r$, the importance of its text feature $i$ is equal to the weight $w_i^r$.

## 4. Experiments

We use the Yelp review dataset to illustrate our results. The training data set had a total of 20,000 reviews made up of 10,020 positive and 9980 negative ones. The testing data (out-of-sample data set) had 10,000 reviews, a roughly balanced combination of 4999 positive and 5001 negative ones. For modeling purpose, each review is padded/cut to a length of 800 tokens.

### 4.1. Yelp Reviews

We first examine the impact of the complexity of filters and ReLU classifier on the model performance. The filter complexity is defined by the number of filters. Specifically, we have three different kinds of filters (i.e. three filter sizes), which correspond to uni-grams, bi-grams and tri-grams in the input text. The complexity is defined by the number of different filters $n_f$ for each n-gram. As a result, the total number of filters is $3 \times n_f$. The complexity of the ReLU classifier is defined as the number of hidden neurons $n_h$ in the two-layer ReLU DNN. As this is a balanced binary classification problem, we choose accuracy and AUC as metrics of model performance. Figure 3 shows the model performances under different complexities.

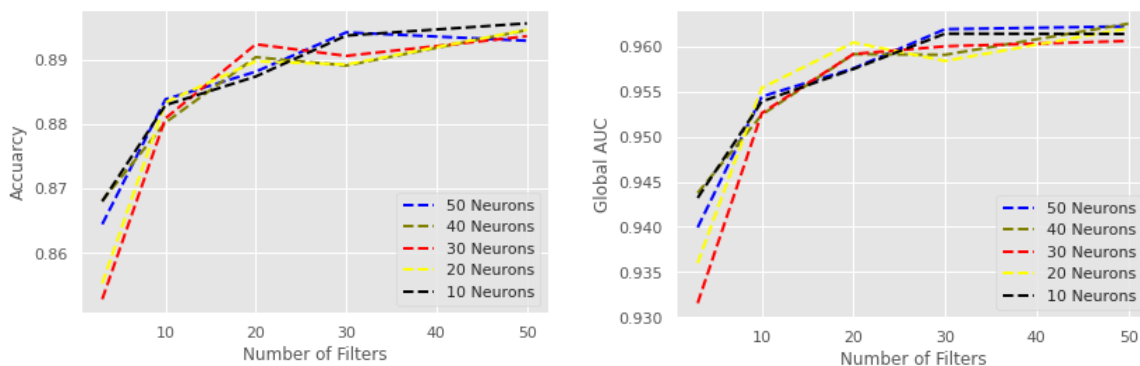

Figure 3: Model performances under different complexities. Specifically, the number of filters per size $n_f$ is chosen among {3,10,20,30,40,50}, the number of hidden neurons $n_h$ is chosen among {10,20,30,40,50}.

From the plots, we can see that as $n_f$, number of filters per filter size, increases from 3 to 50, there is a significant increase in model performance in terms of accuracy and AUC. The increase is most significant for filter size between $n_f = 3$ and $n_f = 10$, followed by the filter size between $n_f = 10$ and $n_f = 20$. As $n_f$ increases further, the additional increase in model performance flattens out. On the other hand, we observe variations in model performance with different ReLU classifier complexities in terms of number of hidden neurons (different curves in Figure 3). But the variation is not consistent across different levels of filter complexity. Therefore, in the Yelp Reviews classification problem, we conclude that the filter complexity is more important than the number of neurons in the ReLU DNN classifier part of the model.

For the following set of experiments, we fix 50 filters per size, which results in a total of $3 \times n_f = 150$ filters in the CNN feature engineering layer. We use $n_h = 30$ as the pervious result shows that the model with 30 hidden neurons has comparable performance with other choices of $n_h$. We apply L1 regularization on the ReLU classifier for the purpose of model simplification. From previous experiments, we concluded that the feature extraction layer contributes more towards model performance than the classification layer. This conclusion is further verified by additional experiments comparing L1 regularization on the max-pooling output layer with L1 on the hidden neurons in linear layer. Figure 4 shows the changes in performance and total LLMs with respect to the changes in the regularization parameter. Both plots show us that accuracy and number of LLMs drop as the L1 penalty increases. However, applying L1 on the hidden neurons results in sharp reduction in the number of LLMs while maintaining a high model performance. In fact, when $\lambda$ is set to be 0, meaning no penalty at all, the test accuracy is 0.907. As the L1 penalty is added and increased, we first observe small fluctuations in model accuracy. At the same time, the number of LLMs decreases significantly even with a small penalty. When $\lambda = 0$, total number of LLMs is 621. But for a penalty $\lambda = 0.01$, the number decreases to 25. Since the complexity of the interpretation of the CNN model is tied to the number of LLMs, optimizing the value of $\lambda$ is key to provide a simple and interpretable model and can help in smaller number of regions and interpretations.

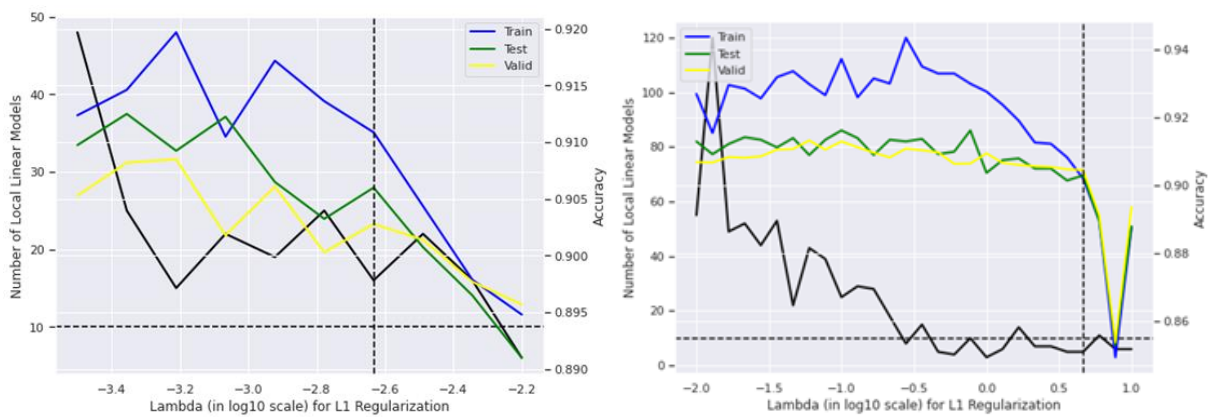

*Figure 4 Relationship of model performance and L1 regularization on different model layers. The left plot shows the results for regularizing input-to-hidden layer, and the right plot shows the results for regularizing the hidden-to-output layer.*

The entire training data set is split into 80% training and 20% validation data. The hyperparameter $\lambda$ is selected by optimizing the performance on the validation set when the number of LLMs are below 10. Therefore, we pick $\lambda = 1.2915$.

Finally, we provide region-wise interpretation by explaining the local linear models. With $\lambda = 1.2915$, the CNN model has test data accuracy of 0.9067 based on six regions (partitions) and associated LLMs. Table 1 shows the number of test samples in each local regions. We see that half of the regions have only 1 sample. Therefore, we use the top three regions with multiple observations to illustrate interpretability. Figure 5 shows the plots of local linear model weights. We can see that model weights follow similar pattern across different local regions but vary across different features. The filters that have top highest local linear weights that contributes largely to the positive prediction are mostly the same among all regions. The same is true for filters that have extremely low weights that contributes largely to the negative prediction.

| Region | Count | Response Mean | Response Std. | Local AUC | Global AUC | Local Accuracy | Global Accuracy | Local F1 | Global F1 |
|---|---|---|---|---|---|---|---|---|---|
| 1 | 4755 | 0.435121 | 0.495773 | 0.899892 | 0.970257 | 0.818717 | 0.9067 | 0.798974 | 0.908323 |
| 2 | 2957 | 0.983429 | 0.127657 | 0.76664 | 0.964331 | 0.983429 | 0.6858 | 0.991645 | 0.760555 |
| 3 | 2285 | 0.009190 | 0.095425 | 0.886968 | 0.966944 | 0.990372 | 0.812 | 0 | 0.772507 |
| 4 | 1 | 0 | 0 | N/A | 0.966955 | N/A | 0.8123 | N/A | 0.772953 |
| 5 | 1 | 1 | 0 | N/A | 0.964313 | N/A | 0.685 | N/A | 0.760091 |
| 6 | 1 | 0 | 0 | N/A | 0.966937 | N/A | 0.8119 | N/A | 0.772359 |

*Table 1 Region information of the ReLU DNN classifier layer of the trained CNN model.*

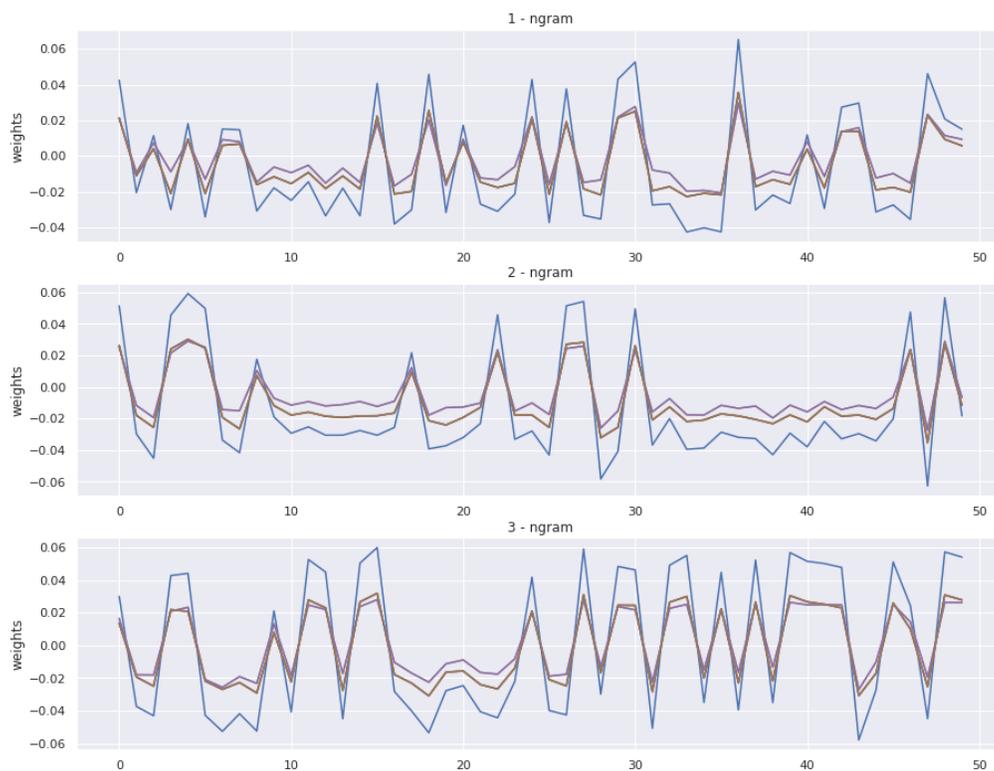

*Figure 5: Local linear weights in each region categorized by filter sizes. As the parameters of some regions are too similar, we can only see 3 distinguished lines in the plot.*

Our model interpretation will focus on the corresponding top positive/negative n-grams. Table 2-5 show the interpretation results in the top 3 local regions. The top filters listed are selected and ranked by their local linear weights (highest to lowest for positive samples and lowest to highest for negative samples) in the corresponding local linear regions. The local n-grams corresponding to each filter are identified by the max-pooling function. As explained in Section 2.1 Structure of ), the submatrix of the embedding matrix, which has the maximum value after the convolution operation, is selected and mapped back to the n-gram in the original text input.

Region 1 is mixed with positive and negative samples, region 2 is a positive region with a response mean of 0.983429, while region 3 is a negative one with a response mean of 0.00919. We show the samples with top five highest prediction scores as well as top 5 lowest prediction scores for region 1 (Table 2 and Table 3). We show the samples with top five highest prediction scores for region 2 (Table 4), and the samples with top five lowest prediction scores for region3 (Table 5). The table rows are ordered by the sample prediction scores. As the weights of the LLM directly reflect the importance of the corresponding filters, we order the table columns by the LLM's filter weights.

For positive samples, we observe an overlap in the important filters between region 1 and region 2, which includes filter 36, 115, 54, 127, 148, 139 and 98. On the other hand, for negative samples, we find that important filters from both region 1 and region 3 contains filter 97, 78, 143, 118, 106, 108, and 131. This finding indicates that additional merging algorithm might further simplify the model.

| Sample ID | Label | Predict | Filter 36 | Filter 115 | Filter 54 | Filter 127 | Filter 148 | Filter 139 | Filter 98 | Filter 133 |
|---|---|---|---|---|---|---|---|---|---|---|
| Weight ($\beta$) | | | 0.065257 | 0.059808 | 0.059202 | 0.058964 | 0.057134 | 0.056748 | 0.056561 | 0.054996 |
| 4194 | 1 | 0.9928 | seasonal | offers more scrumptious | highly recommended | chicken with vegetables | is a good | scrumptious bread and | chocolate waffle | chicken bacon and |
| 8213 | 1 | 0.9904 | night | watching a DVD | I recommend | pizza with sausage | also very entertaining | soft warm bread | cheese pizza | cheese pizza cheese |
| 9286 | 1 | 0.9903 | week | fresh vegetables my | they recommended | nobu seared scallops | still a nice | sautéed fresh vegetables | best lava | lobster cream sauce |
| 5705 | 1 | 0.9899 | evening | fresh and soft | I recommend | combo with hot | is a premium | overpowering fennel seed | Italian sausage | fennel seed flavor |
| 3549 | 1 | 0.9886 | specials | fresh and easy | carefully selected | debates about fresh | now an avid | reasonable prices selfcheck | inexpensive chocolate | delicious oatmeal cookies |

*Table 2 Interpretations of top 5 positive samples in local linear region 1*

| Sample ID | Label | Predict | Filter 97 | Filter 78 | Filter 143 | Filter 118 | Filter 106 | Filter 108 | Filter 131 | Filter 52 |
|---|---|---|---|---|---|---|---|---|---|---|
| Weight ($\beta$) | | | −0.062822 | −0.058329 | −0.057910 | −0.053477 | −0.052625 | −0.052448 | −0.050766 | −0.045211 |
| 652 | 0 | 0.0034 | tremendously disappointing | tremendously disappointing | mushy flavorless rice | was pretty tasteless | out I asked | atrocious mushy flavorless | bad news disappoint | creamed corn |
| 5308 | 0 | 0.0038 | an inaccurate tirade defense | food lackluster | senseless tirade defense | is an inaccurate | senseless tirade defense | appalling and the | our experience nothing | amok fish |
| 4405 | 0 | 0.0042 | extremely salty | extremely salty | filet with cheese | is getting annoying | and I ordered | nasty looking roasted | the waiter did | glorified taco |
| 4211 | 0 | 0.0052 | was inedible | very bad | comped our meal | was very bad | years I ordered | horrible first I | my review because | hubby took |

| Sample ID | label | Predict | | | | | | | | |
|---|---|---|---|---|---|---|---|---|---|---|
| 9000 | 0 | 0.0053 | was bland | really confused | drip of fluids | am really confused | before we chose | violent and uncontrollable | acute food poisoning | vomit up |

*Table 3: Interpretations of top 5 negative samples in local linear region 1*

| Sample ID | label | Predict | Filter 36 | Filter 54 | Filter 98 | Filter 127 | Filter 115 | Filter 30 | Filter 148 | Filter 139 |
|---|---|---|---|---|---|---|---|---|---|---|
| | Weight ($\beta$) | | 0.029656 | 0.029026 | 0.028967 | 0.028002 | 0.027942 | 0.027668 | 0.026322 | 0.026318 |
| 322 | 1 | 0.9997 | valley | highly recommended | favorite sushi | tempura by far | spicy rocking shrimp | best | is amazing great | highly recommended <PAD> |
| 4687 | 1 | 0.9996 | deer | highly recommend | best bread | bacon bits lettuce | fresh sour dough | best | has a nice | bread fresh sour |
| 9539 | 1 | 0.9996 | morning | beautifully roasted | fabulous drinks | coffee beautifully roasted | roasted and smooth | fabulous | is a little | beautifully roasted and |
| 162 | 1 | 0.9995 | town | highly recommend | great prices | usually super fresh | fresh and great | best | is always helpful | always fresh and |
| 2584 | 1 | 0.9993 | weekends | amazing I | favorite pizza | pepperoni and ricotta | weekends and dc | amazing | is a little | good <PAD> <PAD> |

*Table 4: Interpretations of top 5 positive samples in local linear region 2*

| Sample ID | label | Predict | Filter 97 | Filter 78 | Filter 118 | Filter 143 | Filter 108 | Filter 131 | Filter 113 | Filter 106 |
|---|---|---|---|---|---|---|---|---|---|---|
| | Weight ($\beta$) | | $-0.035348$ | $-0.032240$ | $-0.030919$ | $-0.030837$ | $-0.029157$ | $-0.028322$ | $-0.027580$ | $-0.026945$ |
| 8329 | 0 | 0.000029 | very disappointing | very disappointing | was very disappointing | refill our drinks | horrific service I | worst experiences ever | food was cold | disappointing we decided |
| 8781 | 0 | 0.000120 | pretty bland | pretty bland | was just ok | gumbo was cold | horrible we asked | horrible we asked | gumbo was cold | back I ordered |
| 5271 | 0 | 0.000136 | and inedible | boobies bad | fries were soggy | disgusting and inedible | disgusting and inedible | average service good | fries were soggy | inedible burger was |
| 7362 | 0 | 0.000156 | very bland | equally terrible | are equally terrible | completely ignoring their | terrible food do | terrible food do | waitresses were always | drinks we waited |
| 1157 | 0 | 0.000219 | pretty bland | pretty mediocre | food was okay | okay not terrible | terrible but nothing | terrible but nothing | salads were boring | <OOV> my judgment |

*Table 5: Interpretations of top 5 negative samples in local linear region 3*

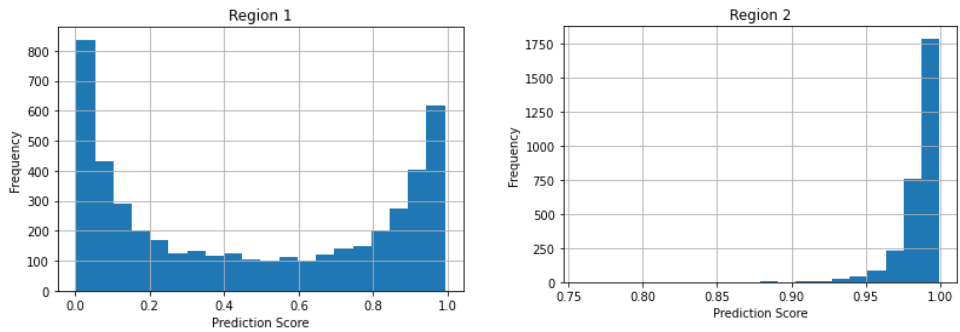

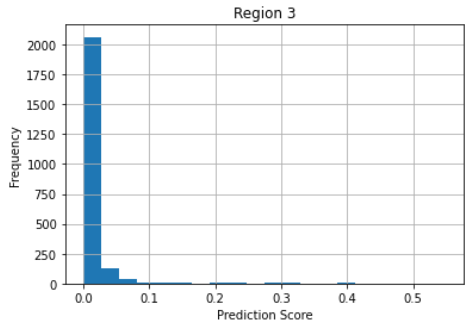

*Figure 6: Distributions of prediction scores in top 3 regions.*

Figure 6 shows the distribution of prediction scores of the samples in the top three regions. The distribution for the second and third region is skewed and imbalanced. The scores for Region 2 range from 0.93 to 1.0, indicating all positive predicted reviews. On the other hand, the scores for Region 3 range from 0 to 0.08, indicating negative predicted reviews. The scores for Region 1, however, range from 0 to 1 covering the full range of positive and negative reviews. Given that region 2 and 3 are skewed towards either positive or negative reviews, the local linear model within each region can potentially be merged. Therefore, we apply the Aletheia merging algorithm, and the local linear regions are finally merged to 1 linear region. The interpretation of the model is given by Table 6-7. It is easy to see that the n-grams selected by each filter express extremely similar meanings and structures. This shows us that top important filters take the role of extracting some specific semantic features that are essential to a document's sentiment. For example, filters that selects n-grams like "highly recommended", "is amazing", and "friendly" contribute mostly to the positive decisions, and filters that selects n-grams like "disappointing", "poor service", and "rude" contribute mostly to the negative decisions.

The results of applying Aletheia's merging algorithm also indicate that the ReLU DNN can be replaced by a single linear layer for the Yelp dataset. To verify the above finding, we refit a CNN with the same feature extraction part followed by a single linear layer classification (Logistic classifier). This model has an accuracy of 90.84% for the Yelp data.

| Sample ID | label | Predict | Filter 77 | Filter 139 | Filter 75 | Filter 36 | Filter 127 | Filter 0 | Filter 130 | Filter 142 |
|---|---|---|---|---|---|---|---|---|---|---|
| **Weight ($\beta$)** | | | 0.158731 | 0.152539 | 0.149857 | 0.143914 | 0.140116 | 0.135715 | 0.132569 | 0.132463 |
| 974 | 1 | 1.0 | spicy rocking | highly recommended <PAD> | is amazing | valley | tempura by far | spicy | roll is amazing | favorite sushi places |
| 503 | 1 | 1.0 | weekends and | good <PAD> <PAD> | is amazing | weekends | pepperoni and ricotta | seated | pepperoni and ricotta | favorite pizza place |
| 8904 | 1 | 1.0 | <OOV> and | beautifully roasted and | any of | morning | coffee beautifully roasted | patio | love their lattes | favorite is the |
| 7981 | 1 | 1.0 | friends and | burrito outstanding blueberry | was great | morning | burrito outstanding blueberry | friendly | friends and guest | niece from los |
| 3281 | 1 | 1.0 | facials and | the fresh apples | is excellent | afternoon | facials and massages | relaxing | facials and massages | best spa in |

*Table 6 Interpretations of top 5 positive samples with positive filter for merged model*

| Sample ID | label | Predict | Filter 51 | Filter 13 | Filter 39 | Filter 143 | Filter 131 | Filter 83 | Filter 126 | Filter 108 |
|---|---|---|---|---|---|---|---|---|---|---|

| | Weight (β) | | −0.293268 | −0.226879 | −0.205832 | −0.182524 | −0.149415 | −0.149408 | −0.136672 | −0.132407 |
|---|---|---|---|---|---|---|---|---|---|---|
| **3406** | 0 | 5.68e-09 | not impressed | rude | horrific | messed up three | horrific extremely slow | extremely slow | the manager accused | horrific extremely slow |
| **9272** | 0 | 1.03e-08 | not happy | school | worst | tasteless pasta dishes | worst ever our | and paid | worst ever our | cold and tasteless |
| **1633** | 0 | 1.10e-08 | not offer | high | worst | crappy employees took | worst experiences we | very rude | worst experiences we | terrible experience crappy |
| **5613** | 0 | 1.20e-08 | it medium | rude | terrible | filet focaccia sandwich | their menu they | was bad | bad the food | terrible and the |
| **1320** | 0 | 1.96e-08 | no interest | rude | horrible | yeah and walked | horrible customer service | horrible customer | horrible customer service | horrible customer service |

*Table 7 Interpretations of top 5 positive samples with positive filter for merged model*

## 4.2. Conclusions

From the experiment of Yelp Reviews data set, we have shown that:

- Feature extraction layer contributes more to the model performance than the ReLU classification layer. Therefore, increasing the number of filters while simplifying the ReLU classifier can help in building a simple but well-performed CNN text classification model;
- L1 regularized ReLU classifier helps in easier self-interpretation of the model as it results in fewer number of partitions and LLMs. The merging algorithm further simplifies the ReLU classifier when redundant LLMs exists;
- For this dataset, an important finding is that a CNN text classification model with convolutional filters performs well with the simple Logistic classifier after max-pooling;
- The interesting feature engineering property of convolutional layer and local linearity property of ReLU DNN classifier combined together provides us easy and intuitive interpretation of the CNN model;
- More specifically, we are able to identify top important filters corresponding to positive/negative sentiment. Furthermore, we can get direct insight into what n-grams type and structure the top important filters tend to select.

## 5. Summary

In the paper, we proposed a novel technique towards building simplified self-interpretable NLP models based on convolutional neural networks (CNNs). The technique consists of three basic steps: model building, model simplification and model interpretation. The methods proposed yield a model that achieves similar performance as other complex models with qualitatively better and easier interpretations. Through an experimental study, we compare the effect of feature engineering layer and ReLU DNN layer on model performance. Furthermore, we evaluated the influences of L1 regularization on both model performance and model interpretation complexity. We introduced using merging algorithm in Aletheia to further simplify the classifier layer of CNN. Finally, we obtained region-wise top important filters and provided interpretation for the model by examining the filter n-grams of the samples within that region. The interpretation based on merged model is also presented by analyzing the top positive and top negative examples. An important finding through the model simplification experiment using

Yelp data set is that a CNN text classification model that has a simple Logistic classifier performs well for this specific data set.

Future research is needed to study model simplification to eliminate unnecessary local linear models. In addition, more work needs to be done to examine how the ReLU classifier may interact with the feature engineering of CNN model. We can also consider transfer learning by selecting top importance filters and feed them into downstream classifiers. Another potential usage of the technique can be model diagnostics for those miss-classified samples by identifying top important filters and n-grams.